# Part-of-Speech Tagging for Code-mixed Indian Social Media Text at ICON 2015

Kamal Sarkar
Computer Science & Engineering Dept.
Jadavpur University
Kolkata-700032, India
jukamal2001@yahoo.com

## ABSTRACT
This paper discusses the experiments carried out by us at Jadavpur University as part of the participation in ICON 2015 task: *POS Tagging for Code-mixed Indian Social Media Text*. The tool that we have developed for the task is based on Trigram Hidden Markov Model that utilizes information from dictionary as well as some other word level features to enhance the observation probabilities of the known tokens as well as unknown tokens. We submitted runs for Bengali-English, Hindi-English and Tamil-English Language pairs. Our system has been trained and tested on the datasets released for ICON 2015 shared task: *POS Tagging For Code-mixed Indian Social Media Text*. In constrained mode, our system obtains average overall accuracy (averaged over all three language pairs) of 75.60% which is very close to other participating two systems (76.79% for IIITH and 75.79% for AMRITA_CEN) ranked higher than our system. In unconstrained mode, our system obtains average overall accuracy of 70.65% which is also close to the system (72.85% for AMRITA_CEN) which obtains the highest average overall accuracy.

## 1. INTRODUCTION
Part-of-Speech (POS) tagging is the task of assigning grammatical categories (noun, verb, adjective etc.) to words in a natural language sentence [1]. POS tagging can be used in various NLP (Natural Language Processing) applications. The interest in applying NLP methods for analyzing non-standardized texts, such as social media texts, rapidly is growing [2], because the automatic analysis of social media texts is one of essential requirements for the task of sentiment analysis [3]. Since social media texts contain blog comments or chat messages, it differs from standardized texts in the word usage but also in their grammatical structure. This creates the need for adapting NLP methods to analyzing social media text and in particular, for the adaption of POS tagging methods to such text types. Most state-of-the art taggers have been developed for standardized texts.

This paper presents a description of HMM (Hidden Markov Model) based system for POS tagging from Social Media Text in Indian Languages.

The ICON 2015 shared task: *POS Tagging For Code-mixed Indian Social Media Text* is defined in this year to build the POS tagger systems for code mixed Indian social media text - Bengali-English, Hindi-English and Tamil-English language pairs for which training data and test data were provided. Data set for a language pair contains the social media text written in the languages of the concerned pair. For example, for Bengali-English language pair, data set contains the social media text written in English and Hindi. We have participated for all three language pairs.

POS Tagger can be developed using both linguistic models and stochastic models. The earliest works on POS tagging [4][5][6] use supervised learning methods.

Some research work has already done for developing POS tagger for standard texts in Indian languages [7].

Dandapat et. al [8].presents HMM and Maximum Entropy (ME) based approaches for Bengali POS tagging.

Ekbal et. al. [9] presented a POS tagger for Bengali language using Conditional Random Fields (CRF). They also discussed another machine learning based POS tagger using SVM algorithm in [10].

An unsupervised Parts-of-Speech Tagger for the Bangla language was proposed by Ali et.al. in [11]. Chakrabarti et.al.[12] has proposed a Layered Parts of Speech Tagging for Bangla.

A detailed survey on POS tagging for other Indian languages has been presented in [13][14].

A few attempts have also been made for developing POS tagger for code mixed Indian social media text. A POS Tagging System of English-Hindi Code-Mixed Social Media Content has been presented in [15]. A POS tagging system for Indian Social Media Text on Twitter has been presented in [16].

## 2. PREPARATION OF TRAINING DATA

The training data released for the ICON 2015 shared task contains three files: one file for Bengali-English Language pair, one file for Hindi-English language pair and one file for Tamil-English language pair. Each line in a file contains tokens in the languages of concerned pair, Language tag and Part-of-Speech tag. The participants are instructed to produce the output in the same format after testing the system on the test data where the test data contains per line a tab separated token and the corresponding language tag. Our system uses a training file for a language pair and converts each sentence into a sequence of pairs of token and tag where each token in this new format is formed by combining the source token and some other information such as language tag. The detailed of this format is discussed in the later sections.

## 3. HMM MODEL FOR POS TAGGING

A POS tagger based on Hidden Markov Model (HMM) finds the best sequence of POS tags $t_1^n$ that is optimal for a given observation sequence $o_1^n$. The tagging problem becomes equivalent to searching for $\arg\max_{t_1^n} P(o_1^n | t_1^n) P(t_1^n)$ (by the application of Bayes' law), that is, we need to compute:

$$\hat{t}_1^n = \arg\max_{t_1^n} P(o_1^n | t_1^n) P(t_1^n) \quad (1).$$

Where $t_1^n$ is a tag sequence and $o_1^n$ is an observation sequence, $P(t_1^n)$ is the prior probability of the tag sequence and $P(o_1^n | t_1^n)$ is the likelihood of the word sequence.

In general, HMM based POS tagging use words in a sentence as an observation sequence [1] [7]. But, we use some additional information such as language tag for disambiguating each token in text. We also use some other information such as whether the token contains any hash tag or not. We use this information in a form of meta tag (details are presented in the subsequent sections). We use a small dictionary of words which contains words with its broad POS categories. If any token is found in the dictionary, we use the broad POS tag as some additional information which we combines with the observation token (details are presented in the subsequent sections).

Unlike the traditional HMM based POS tagging system, to use this additional information for POS tagging task, we consider a triplet as an observation symbol: <word, meta-tag, Language tag >. This is a pseudo token used as an observed symbol, that is, for a sentence of *n* words, the corresponding observation sequence will be as follows:

(<word$_1$, meta-tag$_1$, L-tag$_1$, >, <word$_2$, meta-tag$_2$, L-tag$_2$>, <word$_3$, meta-tag$_3$, L-tag$_3$ >, .........., <word$_n$, meta-tag$_n$, L-tag$_n$,>) . Here an observation symbol $o_i$ corresponds to <word$_i$, meta-tag$_i$, L-tag$_i$, > and L-tag is the language tag and meta-tag is decided based on the additional information (e.g. Hash tag).

Since Equation (1) is too hard to compute directly, HMM taggers follows Markov assumption according to which the probability of a tag is dependent only on short memory (a small, fixed number of previous tags). For example, a bigram tagger considers that the probability of a tag depends only on the previous tag

For our proposed trigram model, the probability of a tag depends on two previous tags and thus $P(t_1^n)$ is computed as:

$$P(t_1^n) \approx \prod_{i=1}^{n} P(t_i | t_{i-1}, t_{i-2}) \quad (2)$$

Depending on the assumption that the probability of a word appearing is dependent only on its own tag, $P(o_1^n | t_1^n)$ can be simplified to:

$$P(o_1^n | t_1^n) \approx \prod_{i=1}^{n} P(o_i | t_i)$$

(3)

Plugging the above mentioned two equations (2) and (3) into (1) results in the following equation by which a bigram tagger estimates the most probable tag sequence:

$$\hat{t}_1^n = \arg\max_{t_1^n} P(t_1^n | o_1^n) P(t_1^n) \approx \arg\max_{t_1^n} \prod_{i=1}^{n} P(o_i | t_i) P(t_i | t_{i-1}) \quad (4)$$

Where: the tag transition probabilities, $P(t_i | t_{i-1})$, represent the probability of a tag given the previous tag. $P(o_i | t_i)$ represents the probability of an observed symbol given a tag.

Considering a special tag $t_{n+1}$ to indicate the end sentence boundary and two special tags $t_{-1}$ and $t_0$ at the starting boundary of the sentence and adding these three special tags to the tag set [4], gives the following equation for POS tagging:

$$\hat{t}_1^n = \arg\max_{t_1^n} P(t_1^n | o_1^n) P(t_1^n) \approx$$

$$\arg\max_{t_1^n} [\prod_{i=1}^{n} P(o_i | t_i) P(t_i | t_{i-1}, t_{i-2})] P(t_{n+1} | t_n) \quad (5)$$

The equation (5) is still computationally expensive because we need to consider all possible tag sequence of length *n*. So, dynamic programming approach is used to compute the equation (5).

At the training phase of HMM based POS tagging, observation probability matrix and tag transition probability matrix are created. A general Architecture of our developed POS tagger is shown in Figure 1.

As we can see from the equation (4), to find the most likely tag sequence for an observation sequence, we need to compute two kinds of probabilities: tag transition probabilities and word likelihoods or observation probabilities.

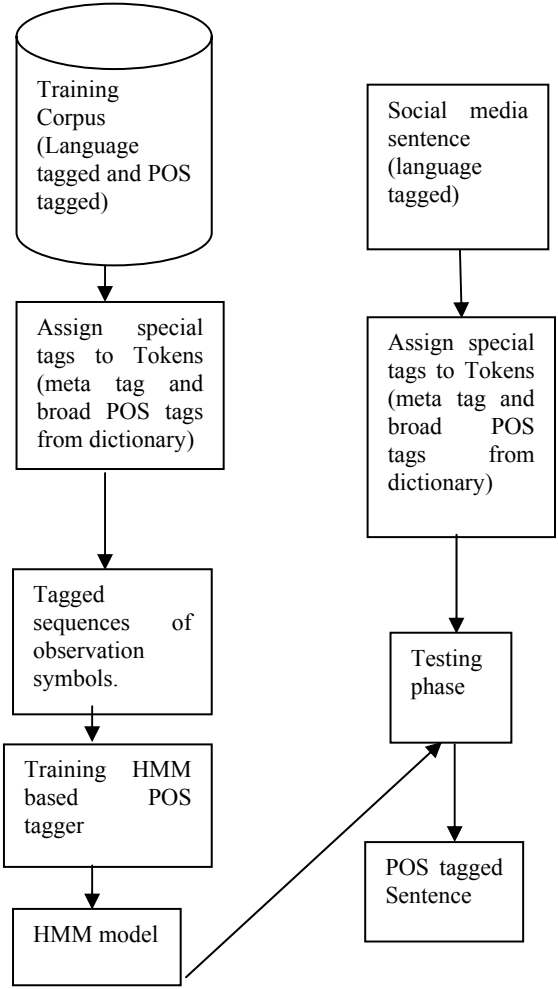

**Figure 1. Architecture for our developed HMM based POS tagging system**

Our developed trigram HMM tagger requires to compute tag trigram probability, $P(t_i | t_{i-1}, t_{i-2})$, which is computed by the maximum likelihood estimate from tag trigram counts. To overcome the data sparseness problem, tag trigram probability is smoothed using deleted interpolation technique [17][4] which uses the maximum likelihood estimates from counts for tag trigram, tag bigram and tag unigram.

The observation probability of a observed triplet <word, meta-tag, L-tag >, which is the observed symbol in our case, is computed using the following equation [1][17].

$$P(o|t) = \frac{C(o,t)}{C(o)} \qquad (7)$$

## 3.1 Viterbi Decoding

We have used Viterbi algorithm to find the best hidden state sequence given an input HMM and a sequence of observation symbols.

The Viterbi algorithm is a standard application of the classic dynamic programming algorithm [18].

Given a tag transition probability matrix and the observation probability matrix, Viterbi decoding (used at the testing phase) accepts a sentence from code mixed social media text and finds the most likely tag sequence for the test sentence which is also L-tagged and Meta tagged. Here a sentence is submitted to the viterbi as the observation sequence of triplets:

(<$word_1$, meta-$tag_1$, L-$tag_1$>, <$word_2$, meta-$tag_2$, L-$tag_2$>, <$word_3$, meta-$tag_3$, L-$tag_3$ >, .........., <$word_n$, meta-$tag_n$, L-$tag_n$ >) . Here an observation symbol $o_i$ corresponds to <$word_i$, meta-$tag_i$, L-$tag_i$,> and L-tag is a language tag and Meta tag is determined based on the dictionary information and Hash tag feature.

After assigning the tag sequence to the observation sequence as mentioned above, L-tag and meta-tag information are removed from the output and thus the output for an input sentence is converted to a POS-tagged sentence.

One of the important problems to apply Viterbi decoding algorithm is how to handle unknown triplets in the input. The unknown triplets are triplets which are not present in the training set and hence their observation probabilities are not known. To handle this problem, we estimate the observation probability of an unknown one by analyzing L-tag, meta-tag and the suffix of the word associated with the corresponding the triplet. We estimate the observation probability of an unknown observed triplet in the following ways:

The observation probabilities of unknown triplet < word, meta-tag, L-tag> corresponding to a word in the input sentence are decided according to the suffix of a pseudo word formed by adding L-tag and meta-tag to the end of the word. We find the observation probabilities of such unknown pseudo words using suffix analysis [17][4]. of all rare pseudo words (frequency <=2) in the training corpus for the concerned language pairs.

## 4. SPECIAL TAGS

### 4.1 Meta Tag

Each token has some properties by which one token differs from another. For example, a token may contain "Hash tag" which is frequent in the social media text.

    Meta-tag="YYYY"(default)
    *if the first character of the token is a Hash symbol (#) then*
    metatag = "HB"

*else if the hash tag is present in any other position of a token*
*metatag = "HE"*
*End If*

### 4.2 Dictionary

In earlier sections, we have mentioned that we have used some dictionary information as the meta-tag also. A meta-tag is set to the value of broad POS tag for a token after matching it with the dictionary words and retrieving the corresponding broad POS tag found in the dictionary. The description of dictionary is shown in Table 1.

**Table 1: Description of Dictionaries**

| Language Pair | Broad POS categories | Number of entries in the dictionary(tokens are not normalized) |
|---|---|---|
| Bengali-English | Pronoun, | 192 |
| | verb | 1791 |
| | and conjunction | 60 |
| Hindi-English | Pronoun | 274 |
| | verb | 1851 |
| | conjunction | 56 |
| Tamil-English | Pronoun | 203 |
| | Verb | 1633 |
| | Conjunction | 56 |

We follow the following rules for assigning to the token this type of broad POS tag extracted from the dictionary:

*If raw token is found in the dictionary and the broad POS tag of the concerned token is XXXX then*

　*meta-tag ="XXXX"*

*end if*

Since we have used only verb, pronoun and conjunctions in the dictionaries, XXXX can take one three values: VERB, PNON and CONJ.

## 5. EVALUATION AND RESULTS

We train separately our developed POS tagger based on the training data and tune the parameters of our system on the training data for the respective language pair. After learning the tuning parameters, we test our system on the test data for the concerned language pair. The description of the data for three language pairs is shown in the Table2

Our developed POS system has been evaluated using the traditional accuracy measure. For training, tuning and testing our system, we have used the datasets for three different language pairs: Bengali-English, Hindi-English and Tamil-English, released by the organizers of ICON 2015 shared task: *POS Tagging For Code-mixed Indian Social Media Text*.

**Table2. The description of the data for various language pairs**

| Language | Total of sentences | |
|---|---|---|
| | Training data | Test data |
| Bengali-English | 2837 | 1459 |
| Hindi-English | 729 | 377 |
| Tamil-English | 639 | 279 |

The organizers of the shared *task* released the data in two phases: in the first phase, training data is released where training data was language tagged and POS tagged. In the second phase, the test data is released where test data was only language tagged. The contestants are instructed to assign POS tags to the sentences in the test file using their developed systems. The tagged test files for test data sets were finally sent to the organizers for evaluation. The organizers evaluate the different runs submitted by the various teams and send the official results to the participating teams. A total of 10 teams submitted their runs for this contest. For each language pair the contests were done in two different modes: Constrained mode and unconstrained mode. In contrained mode, the participant team is only allowed to use the training corpus. No external resource is allowed. In unconstrained mode, the participant team is allowed to use any external resources (POS tagger, NER, Parser, and additional data) to train their system.

In constrained mode, we have not used any dictionary and only Hash tag has been used as the meta-tag. In unconstrained mode, we have used a small dictionary as mentioned in Table 1 and Hash tag has been used as the meta-tag.

The results obtained by our system (team code: KS_JU) have been shown in the tables 3 to 8. The results obtained by other participating systems have also been shown in the tables. The second row of the each table shows the overall accuracy obtained by the various systems participated in the contest.

We have also evaluated the system based on its consistency across the languages in constrained and unconstrained mode.

Average overall accuracy is computed by taking the average of overall accuracy of the system obtained for all three language pairs in a particular mode.

In constrained mode, our system obtains average overall accuracy (averaged over all three language pairs) of 75.60% which is very close to other participating two systems (76.79% for IIITH and 75.79% for AMRITA_CEN) ranked higher than our system. In unconstrained mode, our system obtains average overall accuracy of 70.65% which is also close to the system (72.85% for AMRITA_CEN) which obtains the highest average overall accuracy.

**Table 3.** Official results (Bengali-Constrained mode) obtained by the various systems participated in ICON 2015 shared task: *POS Tagging For Code-mixed Indian Social Media Text*

| POS/Categorical | IIITH | AMRITA_CEN | KS_JU | CDACMUMBAI | DD_JU | SN_JU | Amrita |
|---|---|---|---|---|---|---|---|
| Overall | 79.84% | 78.50% | **78.42%** | 75.46% | 75.22% | 72.64% | 10.13% |
| E | 97.11% | 94.22% | 97.11% | 97.11% | 95.95% | 97.11% | 0.00% |
| @ | 100.00% | 93.33% | 93.33% | 93.33% | 86.67% | 86.67% | 0.00% |
| JJ | 65.25% | 61.12% | 61.92% | 62.72% | 58.19% | 52.46% | 20.51% |
| N_NST | 80.00% | 80.00% | 80.00% | 80.00% | 0.00% | 80.00% | 0.00% |
| DT | 95.90% | 96.29% | 94.92% | 95.51% | 93.75% | 94.73% | 0.00% |
| RD_SYM | 0.00% | 0.00% | 0.00% | 0.00% | 0.00% | 0.00% | 0.00% |
| RB_AMN | 81.48% | 77.78% | 80.79% | 77.31% | 76.16% | 66.20% | 0.00% |
| N_NN | 81.26% | 79.80% | 78.18% | 81.73% | 83.56% | 67.66% | 11.20% |
| U | 100.00% | 13.64% | 81.82% | 100.00% | 0.00% | 100.00% | 0.00% |
| RD_RDF | 47.96% | 40.52% | 42.94% | 39.22% | 36.06% | 33.64% | 0.00% |
| QT_QTF | 48.75% | 55.63% | 57.50% | 56.25% | 53.13% | 50.00% | 0.00% |
| RP_RPD | 71.24% | 74.51% | 76.47% | 69.28% | 49.02% | 77.78% | 0.00% |
| N_NNV | 59.68% | 62.90% | 56.45% | 35.48% | 66.13% | 56.45% | 0.00% |
| V_VM | 79.76% | 81.87% | 78.49% | 80.66% | 74.76% | 71.81% | 0.54% |
| PR_PRQ | 83.93% | 87.50% | 75.00% | 87.50% | 91.07% | 82.14% | 0.00% |
| # | 95.35% | 97.67% | 97.67% | 88.37% | 74.42% | 74.42% | 0.00% |
| PR_PRP | 87.48% | 90.19% | 88.77% | 89.29% | 87.10% | 87.61% | 0.00% |
| N_NNP | 65.46% | 55.47% | 59.52% | 59.81% | 43.08% | 61.55% | 60.68% |
| V_VAUX | 39.08% | 31.03% | 35.06% | 27.59% | 20.69% | 30.46% | 0.00% |
| $ | 64.71% | 69.85% | 61.76% | 61.76% | 41.91% | 44.85% | 0.00% |
| RP_INJ | 53.61% | 50.52% | 60.82% | 54.64% | 26.80% | 49.48% | 0.00% |
| RB_ALC | 54.41% | 70.59% | 58.82% | 63.24% | 75.00% | 54.41% | 0.00% |
| DM_DMD | 71.34% | 72.61% | 74.52% | 70.70% | 78.98% | 76.43% | 0.00% |
| PR_PRF | 55.56% | 77.78% | 44.44% | 55.56% | 77.78% | 66.67% | 0.00% |
| CC | 82.76% | 85.17% | 85.52% | 83.79% | 83.10% | 81.38% | 0.34% |
| DM_DMQ | 50.00% | 50.00% | 50.00% | 50.00% | 0.00% | 50.00% | 0.00% |
| PSP | 87.69% | 89.38% | 92.36% | 90.54% | 87.56% | 89.25% | 3.89% |
| DM_DMR | 0.00% | 0.00% | 0.00% | 0.00% | 0.00% | 0.00% | 0.00% |
| RD_PUNC | 98.79% | 99.11% | 98.46% | 76.74% | 97.67% | 93.57% | 0.51% |
| PR_PRL | 60.00% | 80.00% | 80.00% | 80.00% | 60.00% | 40.00% | 0.00% |

**Table 4.** Official results (Bengali_unconstrained) obtained by the various systems participated in ICON 2015 shared task: *POS Tagging For Code-mixed Indian Social Media Text*

| POS/Categorical | KS_JU | AMRITA_CEN | DD_JU |
|---|---|---|---|
| Overall | **78.29%** | 76.73% | 47.08% |
| E | 58.96% | 94.80% | 95.95% |
| @ | 66.67% | 93.33% | 86.67% |
| JJ | 45.94% | 61.38% | 56.32% |
| N_NST | 50.00% | 80.00% | 0.00% |
| DT | 59.96% | 96.29% | 61.72% |
| RD_SYM | 0.00% | 0.00% | 0.00% |
| RB_AMN | 53.70% | 80.56% | 0.23% |
| N_NN | 57.68% | 76.18% | 44.86% |
| U | 31.82% | 9.09% | 0.00% |
| RD_RDF | 29.74% | 36.62% | 36.06% |
| QT_QTF | 41.25% | 54.37% | 53.13% |
| RP_RPD | 52.94% | 66.67% | 33.33% |
| N_NNV | 33.87% | 59.68% | 48.39% |
| V_VM | 48.37% | 79.46% | 15.54% |
| PR_PRQ | 48.21% | 89.29% | 91.07% |
| # | 74.42% | 95.35% | 74.42% |
| PR_PRP | 60.52% | 89.03% | 18.32% |
| N_NNP | 38.23% | 50.18% | 42.22% |
| V_VAUX | 16.09% | 35.63% | 20.69% |
| $ | 38.24% | 72.79% | 28.68% |
| RP_INJ | 25.77% | 60.82% | 14.43% |
| RB_ALC | 45.59% | 66.18% | 75.00% |
| DM_DMD | 54.14% | 74.52% | 78.98% |
| PR_PRF | 33.33% | 100.00% | 77.78% |
| CC | 59.66% | 83.79% | 73.79% |
| DM_DMQ | 25.00% | 25.00% | 0.00% |
| PSP | 59.33% | 88.86% | 16.97% |
| DM_DMR | 0.00% | 0.00% | 0.00% |
| RD_PUNC | 64.34% | 98.93% | 97.67% |
| PR_PRL | 40.00% | 80.00% | 60.00% |

**Table 5. Official results (Hindi-constrained) obtained by the various systems participated in ICON 2015 shared task: *POS Tagging For Code-mixed Indian Social Media Text***

| POS/Categorical | KS_JU | AMRITA_CEN | IIITH | DD_JU | CDACMUMBAI | SN_JU | Anuj_IITB | Amrita |
|---|---|---|---|---|---|---|---|---|
| Overall | **77.74%** | 75.58% | 75.04% | 73.16% | 71.11% | 68.85% | 64.52% | 13.45% |
| E | 7.94% | 94.44% | 94.44% | 92.06% | 94.44% | 91.27% | 96.03% | 1.59% |
| @ | 16.67% | 83.33% | 50.00% | 33.33% | 83.33% | 33.33% | 83.33% | 0.00% |
| JJ | 9.93% | 52.23% | 56.40% | 54.10% | 56.55% | 55.68% | 64.60% | 0.86% |
| DT | 15.74% | 93.77% | 92.07% | 90.26% | 90.49% | 91.39% | 86.98% | 0.00% |
| N_NST | 0.00% | 0.00% | 0.00% | 0.00% | 0.00% | 0.00% | 0.00% | 0.00% |
| RB_AMN | 15.58% | 75.88% | 76.42% | 78.32% | 77.78% | 65.18% | 69.65% | 0.00% |
| RD_SYM | 0.00% | 91.67% | 91.67% | 91.67% | 91.67% | 91.67% | 91.67% | 0.00% |
| N_NN | 13.93% | 79.83% | 82.77% | 81.75% | 82.77% | 71.97% | 48.38% | 20.89% |
| U | 0.00% | 12.50% | 62.50% | 0.00% | 62.50% | 93.75% | 93.75% | 0.00% |
| RD_RDF | 0.76% | 4.55% | 3.03% | 3.79% | 3.79% | 4.55% | 3.79% | 0.00% |
| QT_QTF | 0.00% | 10.00% | 10.00% | 10.00% | 10.00% | 10.00% | 20.00% | 0.00% |
| RP_RPD | 0.00% | 0.00% | 0.00% | 27.78% | 0.00% | 5.56% | 5.56% | 0.00% |
| N_NNV | 4.76% | 9.52% | 9.52% | 4.76% | 9.52% | 9.52% | 9.52% | 0.00% |
| RP_INTF | 0.00% | 0.00% | 0.00% | 0.00% | 0.00% | 0.00% | 0.00% | 0.00% |
| V_VM | 16.68% | 83.32% | 81.42% | 84.49% | 82.46% | 74.62% | 52.30% | 56.84% |
| PR_PRQ | 0.00% | 88.89% | 66.67% | 22.22% | 33.33% | 33.33% | 44.44% | 0.00% |
| # | 20.97% | 100.00% | 100.00% | 100.00% | 100.00% | 80.65% | 100.00% | 0.00% |
| RD_UNK | 0.00% | 0.00% | 0.00% | 0.00% | 0.00% | 0.00% | 0.00% | 0.00% |
| PR_PRP | 18.68% | 87.18% | 82.67% | 88.54% | 79.69% | 88.09% | 73.01% | 0.45% |
| N_NNP | 11.99% | 67.54% | 69.30% | 67.84% | 53.22% | 35.38% | 69.88% | 2.63% |
| V_VAUX | 8.98% | 34.04% | 41.13% | 6.38% | 36.41% | 43.26% | 50.35% | 1.65% |
| $ | 9.81% | 69.16% | 65.89% | 36.45% | 57.94% | 37.38% | 57.01% | 0.00% |
| RP_INJ | 4.76% | 61.90% | 55.24% | 43.81% | 54.29% | 43.81% | 47.62% | 0.95% |
| RB_ALC | 0.00% | 6.67% | 6.67% | 0.00% | 6.67% | 0.00% | 6.67% | 0.00% |
| PR_PRF | 0.00% | 0.00% | 0.00% | 0.00% | 0.00% | 0.00% | 0.00% | 0.00% |
| CC | 11.92% | 34.89% | 45.94% | 7.60% | 41.45% | 44.21% | 53.54% | 0.00% |
| PSP | 9.07% | 75.67% | 62.37% | 82.99% | 69.18% | 62.78% | 58.66% | 0.82% |
| ~ | 0.00% | 0.00% | 0.00% | 0.00% | 0.00% | 0.00% | 0.00% | 0.00% |
| RD_PUNC | 18.44% | 98.30% | 97.85% | 95.85% | 70.52% | 85.33% | 96.22% | 14.44% |
| PR_PRL | 1.45% | 0.00% | 1.45% | 0.00% | 1.45% | 0.00% | 5.80% | 0.00% |

**Table 6.** Official results (Hindi-Unconstrained) obtained by the various systems participated in ICON 2015 shared task: *POS Tagging For Code-mixed Indian Social Media Text*

| POS/Categorical | IIITH | KS_JU | AMRITA_CEN | Rudra_IITB | DD_JU | CDACMUMBAI |
|---|---|---|---|---|---|---|
| Overall | 80.68% | **77.60%** | 73.66% | 68.94% | 27.60% | 6.84% |
| E | 98.41% | 7.94% | 93.65% | 96.03% | 92.06% | 5.56% |
| @ | 83.33% | 16.67% | 66.67% | 50.00% | 33.33% | 16.67% |
| JJ | 82.88% | 10.36% | 61.73% | 52.37% | 54.82% | 2.45% |
| DT | 93.54% | 15.52% | 94.11% | 87.32% | 76.90% | 2.49% |
| N_NST | 0.00% | 0.00% | 0.00% | 0.00% | 0.00% | 0.00% |
| RB_AMN | 89.70% | 15.58% | 79.27% | 53.66% | 0.27% | 5.15% |
| RD_SYM | 91.67% | 0.00% | 50.00% | 75.00% | 91.67% | 0.00% |
| N_NN | 88.48% | 14.47% | 81.57% | 71.91% | 4.44% | 26.83% |
| U | 62.50% | 0.00% | 37.50% | 93.75% | 0.00% | 0.00% |
| RD_RDF | 3.03% | 0.76% | 8.33% | 2.27% | 3.03% | 0.76% |
| QT_QTF | 10.00% | 0.00% | 10.00% | 10.00% | 10.00% | 0.00% |
| RP_RPD | 16.67% | 0.00% | 44.44% | 11.11% | 0.00% | 0.00% |
| N_NNV | 9.52% | 4.76% | 9.52% | 0.00% | 4.76% | 4.76% |
| RP_INTF | 0.00% | 0.00% | 0.00% | 0.00% | 0.00% | 0.00% |
| V_VM | 86.82% | 15.57% | 88.78% | 75.17% | 3.49% | 5.70% |
| PR_PRQ | 66.67% | 0.00% | 11.11% | 22.22% | 22.22% | 0.00% |
| # | 100.00% | 20.97% | 100.00% | 90.32% | 100.00% | 0.00% |
| RD_UNK | 0.00% | 0.00% | 0.00% | 0.00% | 0.00% | 0.00% |
| PR_PRP | 87.00% | 18.59% | 87.45% | 90.52% | 2.26% | 1.08% |
| N_NNP | 71.64% | 11.99% | 59.94% | 68.71% | 67.84% | 0.29% |
| V_VAUX | 43.03% | 8.04% | 6.62% | 48.46% | 4.02% | 6.62% |
| $ | 68.22% | 10.28% | 66.36% | 48.60% | 23.36% | 0.00% |
| RP_INJ | 74.29% | 4.76% | 63.81% | 54.29% | 30.48% | 9.52% |
| RB_ALC | 0.00% | 0.00% | 0.00% | 0.00% | 0.00% | 0.00% |
| PR_PRF | 0.00% | 0.00% | 0.00% | 0.00% | 0.00% | 0.00% |
| CC | 50.26% | 11.74% | 8.64% | 89.12% | 3.11% | 0.52% |
| PSP | 65.15% | 10.21% | 60.62% | 13.71% | 3.61% | 1.24% |
| ~ | 0.00% | 0.00% | 0.00% | 0.00% | 0.00% | 0.00% |
| RD_PUNC | 98.15% | 18.37% | 99.11% | 97.04% | 95.85% | 5.48% |
| PR_PRL | 1.45% | 1.45% | 1.45% | 0.00% | 0.00% | 0.00% |

**Table 7.** Official results (Tamil_Constrained) obtained by the various systems participated in ICON 2015 shared task: *POS Tagging For Code-mixed Indian Social Media Text*

| POS/Categorical | IIITH | AMRITA_CEN | CDACMUMBAI | KS_JU | DD_JU | SN_JU | Amrita |
|---|---|---|---|---|---|---|---|
| Overall | 75.48% | 73.30% | 71.04% | **70.64%** | 64.83% | 62.44% | 17.07% |
| N_NNP | 100.00% | 99.09% | 80.91% | 99.09% | 98.64% | 69.55% | 8.64% |
| PR_PRP | 80.92% | 69.08% | 77.10% | 71.37% | 81.30% | 66.41% | 3.44% |
| QT_QTO | 55.56% | 100.00% | 62.96% | 81.48% | 96.30% | 70.37% | 0.00% |
| V_VAUX | 0.00% | 0.00% | 0.00% | 0.00% | 0.00% | 27.27% | 0.00% |
| JJ | 69.70% | 52.02% | 64.65% | 64.14% | 61.11% | 56.57% | 3.54% |
| RP_INJ | 0.00% | 0.00% | 0.00% | 0.00% | 25.00% | 25.00% | 0.00% |
| DT | 79.59% | 65.31% | 71.43% | 73.47% | 91.84% | 61.22% | 0.00% |
| RB_AMN | 59.57% | 46.10% | 59.57% | 53.90% | 43.26% | 43.97% | 7.09% |
| N_NN | 76.52% | 77.64% | 75.72% | 72.52% | 60.70% | 64.70% | 16.61% |
| CC | 73.46% | 79.01% | 77.78% | 76.54% | 62.96% | 78.40% | 0.62% |
| PSP | 66.67% | 52.38% | 49.21% | 50.79% | 58.73% | 60.32% | 0.00% |
| V_VM | 76.81% | 84.54% | 71.98% | 69.81% | 57.49% | 61.59% | 56.76% |
| X | 58.06% | 48.39% | 46.77% | 45.16% | 33.87% | 46.77% | 0.00% |
| RD_PUNC | 0.00% | 0.00% | 0.00% | 0.00% | 0.00% | 0.00% | 0.00% |

**Table 8.** Official results (Tamil_unconstrained) obtained by the various systems participated in ICON 2015 shared task: *POS Tagging For Code-mixed Indian Social Media Text*

| POS/Categorical | AMRITA_CEN | KS_JU | CDACMUMBAI | DD_JU |
|---|---|---|---|---|
| Overall | 68.16% | **56.05%** | 48.03% | 44.21% |
| N_NNP | 80.91% | 99.09% | 7.73% | 98.64% |
| PR_PRP | 72.52% | 27.48% | 39.31% | 54.20% |
| QT_QTO | 74.07% | 81.48% | 51.85% | 96.30% |
| V_VAUX | 0.00% | 0.00% | 9.09% | 0.00% |
| JJ | 59.09% | 66.16% | 38.38% | 54.04% |
| RP_INJ | 50.00% | 0.00% | 0.00% | 0.00% |
| DT | 77.55% | 63.27% | 38.78% | 83.67% |
| RB_AMN | 51.77% | 56.03% | 37.59% | 23.40% |
| N_NN | 68.85% | 71.88% | 90.58% | 16.29% |
| CC | 80.86% | 32.10% | 75.31% | 60.49% |
| PSP | 53.97% | 19.05% | 22.22% | 57.14% |
| V_VM | 69.81% | 41.06% | 17.39% | 42.51% |
| X | 40.32% | 43.55% | 40.32% | 30.65% |
| RD_PUNC | 56.25% | 0.00% | 0.00% | 0.00% |

# 6. CONCLUSION

This paper describes a POS tagging system for code mixed social media text in Indian Languages. The features such as dictionary based information and some other word level features have been introduced into the HMM model. The experimental results show that performance of our system is comparable with the best performing systems participated in ICON 2015 task: POS Tagging for Code-mixed Indian Social Media Text. The POS tagging system has been developed using Visual Basic platform so that a suitable user interface can be designed for the novice users. The system has been designed in such a way that only changing the training corpus in a file can make the system portable to other Indian languages.